\pdfoutput=1

\documentclass[11pt]{article}

\usepackage{acl}

\usepackage{times}
\usepackage{latexsym}
\usepackage{CJKutf8}

\usepackage[T1]{fontenc}

\usepackage[utf8]{inputenc}

\usepackage{microtype}

\usepackage{amsmath}
\usepackage{amssymb}
\usepackage{xcolor}
\usepackage{arydshln}
\usepackage{tikz}
\usetikzlibrary{arrows, decorations.text, shapes.geometric, positioning, decorations.pathreplacing, calligraphy}
\usepackage{pgfplots}
\usepackage{pgfmath}
\usepackage{pgffor}
\pgfplotsset{compat=1.17}
\usepackage{subcaption}
\usepackage{mathtools}
\usepackage{multirow}
\usepackage{booktabs}
\usepackage{pifont}
\usepackage{ifthen}

\title{Alibaba-Translate China's Submission for\\WMT 2022 Quality Estimation Shared Task}

\author{
    Keqin Bao$^{1,2}$\thanks{~~Equal contribution. Work was done when Keqin Bao and Yu Wan were interning at DAMO Academy, Alibaba Group.}~~~Yu Wan$^{1,3*}$~~~Dayiheng Liu$^1$~~~Baosong Yang$^1$~~~\textbf{Wenqiang Lei}$^4$\\\textbf{Xiangnan He}$^2$~~~\textbf{Derek F. Wong}$^3$~~~\textbf{Jun Xie}$^1$\\
    $^1$DAMO Academy, Alibaba Group~~~~~~$^2$University of Science and Technology of China\\
    $^3$NLP$^2$CT Lab, University of Macau~~~~~~$^4$National University of Singapore\\
    \vspace{-0.75ex}
    \small{\tt{baokq@mail.ustc.edu.cn~~~nlp2ct.ywan@gmail.com}}\\
    \vspace{-0.75ex}
    \small{\tt{\{liudayiheng.ldyh,yangbaosong.ybs,qingjing.xj\}@alibaba-inc.com}}\\
    \vspace{-0.75ex}
    \small{\tt{wenqianglei@gmail.com}~~~\tt{xiangnanhe@gmail.com}~~~\tt{derekfw@um.edu.mo}}
}

\begin{document}
\maketitle



\begin{abstract}
In this paper, we present our submission to the sentence-level MQM benchmark at Quality Estimation Shared Task, named~\textsc{\textbf{UniTE}} (\textbf{Uni}fied \textbf{T}ranslation \textbf{E}valuation).
Specifically, our systems employ the framework of~\textsc{UniTE}, which combined three types of input formats during training with a pre-trained language model.
First, we apply the pseudo-labeled data examples for the continuously pre-training phase.
Notably, to reduce the gap between pre-training and fine-tuning, we use data pruning and a ranking-based score normalization strategy.
For the fine-tuning phase, we use both Direct Assessment (DA) and Multidimensional Quality Metrics (MQM) data from past years' WMT competitions.
Finally, we collect the source-only evaluation results, and ensemble the predictions generated by two~\textsc{UniTE} models, whose backbones are XLM-R and~\textsc{InfoXLM}, respectively.
Results show that our models reach 1st overall ranking in the Multilingual and English-Russian settings, and 2nd overall ranking in English-German and Chinese-English settings, showing relatively strong performances in this year's quality estimation competition.

\end{abstract}
\section{Introduction}
\label{sec.intro}
Quality Estimation (QE) aims at evaluating machine translation without access to a gold-standard reference translation~\cite{blatz-etal-2004-confidence,specia-etal-2018-findings}.
Different from other evaluation tasks (\textit{e.g.}, metric), QE arranges its process of evaluation via only accessing source input.
As the performance of modern machine translation approaches increase~\cite{vaswani2017attention,lin2022bridging, wei-etal-2022-learning, zhang2022frequency}, the QE systems should better quantify the agreement of cross-lingual semantics on source sentence and translation hypothesis.
The evaluation paradigm of QE shows its own potential for real-world applications~\cite{wang-etal-2021-qemind, park-etal-2021-papagos, specia-etal-2021-findings}.
This paper describes Alibaba Translate China's submission to the sentence-level MQM benchmark at WMT 2022 Quality Estimation Shared Task~\cite{zerva-EtAl:2022:WMT}.

In recent years, pre-trained language models (PLMs) have shown their strong ability on extracting cross-lingual information \cite{conneau-etal-2020-unsupervised, chi-etal-2021-infoxlm}.
To achieve a higher correlation with human ratings on the quality of translation outputs, plenty of trainable model-based QE approaches appear,~\textit{e.g.}, \textsc{COMET-QE}~\cite{rei-etal-2020-comet} and \textsc{QEMinD}~\cite{wang-etal-2021-qemind}.
They both first derive the embeddings assigned with source and hypothesis sentence with given PLM, then predict the overall score based on their embeddings with a followed feedforward network.
Those model-based approaches have greatly facilitated the development of the QE community.
However, those models can only handle source-only input format, which neglects the other two evaluation scenarios, \textit{i.e.}, reference-only and source-reference-combined evaluation.
More importantly, training with multiple input formats can achieve a higher correlation with human assessments than individually training on specific evaluation scenarios~\cite{wan-etal-2021-robleurt,wan-etal-2022-unite}.
Those findings indicate that, the QE and Metric tasks share plenty of knowledge when identifying the quality of translated outputs, and unifying the functionalities of three evaluation scenarios into one model can also enhance the performance of the evaluation model on each scenario.

As a consequence, when building a single model for a sentence-level QE task, we use the pipeline of \textsc{UniTE}~\cite{wan-etal-2022-unite}, which integrates source-only, reference-only, and source-reference-combined translation evaluation ability into one single model.
When collecting the system outputs for WMT 2022 Quality Estimation Shared Task, we employ our~\textsc{UniTE} models to predict the translation quality scores following a source-only setting.
As for the training data, we collect synthetic data examples as supervision for continuous pre-training and apply a dataset pruning strategy to increase the translation quality of the training set.
Also, during fine-tuning our QE model, we use all available Direct Assessment~\citep[DA,][]{bojar-etal-2017-results,ma-etal-2018-results,ma-etal-2019-results,mathur-etal-2020-results} and Multidimensional Quality Metrics datasets~\citep[MQM,][]{freitag-etal-2021-experts,freitag-etal-2021-results} from previous WMT competitions to further improve the performance of our model.
Besides, regarding the applied PLM for~\textsc{UniTE} models, we find that for English-Russian (En-Ru) and Chinese-English (Zh-En) directions, PLM enhanced with cross-lingual alignments ~\citep[\textsc{InfoXLM},][]{chi-etal-2021-infoxlm} can deliver better results than conventional ones~\citep[XLM-R,][]{conneau-etal-2020-unsupervised}.
Moreover, for each subtask including English to German (En-De), En-Ru, Zh-En, and multilingual direction evaluations, we build an ensembled QE system to derive more accurate and convincing results as final predictions.

Our models show impressive performances in all translation directions.
When only considering the primary metric -- Spearman's correlation, we get 2nd, 3rd, and 3rd place in En-Ru, Zh-En, and multilingual direction, respectively.
More notably, when taking all metrics into account, despite the slight decrease in Spearman's correlations, our systems show outstanding overall performance than other systems, achieving 1st place in En-Ru and multilingual, and 2nd in En-De and Zh-En direction.

\section{Method}
As outlined in \S\ref{sec.intro}, we apply the~\textsc{UniTE} framework~\cite{wan-etal-2022-unite} to obtain QE models.
We unify three types of input formats (\textit{i.e.}, source-only, reference-only, and source-reference-combined) into one single model during training. 
While during inference, we only use the source-only paradigm to collect evaluation scores.
In this section, we introduce the applied model architecture (\S\ref{subsec.model_architecture}), synthetic data construction method (\S\ref{subsec.constructing_synthetic_data}), and model training strategy (\S\ref{subsec.training_pipeline}).

\subsection{Model architecture}
\label{subsec.model_architecture}
\paragraph{Input Format}
Following~\newcite{wan-etal-2022-unite}, we design our QE model which is capable of processing ~\textbf{source-only},~\textbf{reference-only}, and~\textbf{source-reference-combined} evaluation scenarios.
Consequently, for the consistency of training across all input formats, we construct the input sequence for source-only, reference-only, and source-reference-combined input formats as follows:
\begin{align}
    \mathbf{x}_{\text{\textsc{Src}}} &= \langle \texttt{s} \rangle \mathbf{h} \langle \texttt{/s} \rangle \langle \texttt{/s} \rangle \mathbf{s} \langle \texttt{/s} \rangle, \\
    \mathbf{x}_{\text{\textsc{Ref}}} &= \langle \texttt{s} \rangle \mathbf{h} \langle \texttt{/s} \rangle \langle \texttt{/s} \rangle \mathbf{r} \langle \texttt{/s} \rangle, \\
    \mathbf{x}_{\text{\textsc{Src+Ref}}} &= \langle \texttt{s} \rangle \mathbf{h} \langle \texttt{/s} \rangle \langle \texttt{/s} \rangle \mathbf{s} \langle \texttt{/s} \rangle \langle \texttt{/s} \rangle \mathbf{r} \langle \texttt{/s} \rangle,
\end{align}
where $\mathbf{h}$, $\mathbf{s}$, and $\mathbf{r}$ represent hypothesis, source, and reference sentence, respectively.
During the pre-training phase, we apply all input formats to enhance the performance of QE models.
Notably, we only use the source-only format setting when fine-tuning on this year's dev set and inferring the test set.

\paragraph{Model Backbone Selection}
The core of quality estimation aims at evaluating the translation quality of output given source sentence.
As the source and hypothesis sentence are from different languages, evaluating the translation quality requires the ability of multilingual processing.
Furthermore, we believe that those PLMs which possess cross-lingual semantic alignments can ease the learning of translation quality evaluation.

Referring to the setting of existing methods~\cite{ranasinghe2020transquest,rei-etal-2020-comet,sellam-etal-2020-bleurt,wan-etal-2022-unite}, they often apply XLM-R~\cite{conneau-etal-2020-unsupervised} as the backbone of evaluation models for better multilingual support.
To testify whether cross-lingual alignments can help the evaluation model training, we further apply~\textsc{InfoXLM}~\cite{chi-etal-2021-infoxlm}, which enhances the XLM-R model with cross-lingual alignments, as the backbone of evaluation models.

\paragraph{Model Training}
For the training dataset including source, reference, and hypothesis sentences, we first equally split all examples into three parts, each of which only serves one input format training.
As to each training example, after concatenating the required input sentences into one sequence and feeding it to PLM, we collect the corresponding representations --  $\mathbf{H}_\textsc{Ref}, \mathbf{H}_\textsc{Src}, \mathbf{H}_\textsc{Src+Ref}$ for each input format, respectively.
After that, we use the output embedding assigned with CLS token $\mathbf{h}$ as the sequence representation.
Finally, a feedforward network takes $\mathbf{h}$ as input and gives a scalar $p$ as a prediction.
Taking $\mathbf{x}_{\textsc{Src}}$ as an example: 
\begin{align}
    \mathbf{H}_\textsc{Src} &= \texttt{PLM}(\mathbf{x}_\textsc{Src}) \in \mathbb{R}^{ (l_h + l_s) \times d}, \\
    \mathbf{h}_\textsc{Src}& = \texttt{CLS}(\mathbf{H}_\textsc{Src}) \in \mathbb{R}^{d}, \\
    p_\textsc{Src} & = \texttt{FeedForward}(\mathbf{h}_\textsc{Src}) \in \mathbb{R}^{1},
\end{align}
where $l_h$ and $l_s$ are the lengths of $\mathbf{h}$ and $\mathbf{s}$, respectively.

For the learning objective, we apply the mean squared error (MSE) as the loss function:
\begin{align}
   \mathcal{L}_\textsc{SRC} = (p_\textsc{SRC} - q) ^ 2,
\end{align}
where $q$ is the given ground-truth score. 
Note that, when training on three input formats, one single step includes three substeps, each of which is arranged on one specific input format.
Besides, the batch size is the same across all input formats to avoid the training imbalance.
During each update, the final learning objective can be written as the sum of losses for each format:
\begin{align}
    \mathcal{L} = \mathcal{L}_\textsc{Ref} + \mathcal{L}_\textsc{Src} + \mathcal{L}_\textsc{Src+Ref}.
\end{align}

\subsection{Constructing Synthetic Data}
\label{subsec.constructing_synthetic_data}
To better enhance the translation evaluation ability of pre-trained models, we first construct synthetic dataset for continuous pre-training~\cite{wan-etal-2022-unite}.
The pipeline for obtaining such dataset consists of the following steps:
1) collecting synthetic data from parallel data provided by the WMT Translation task;
2) labeling samples with a ranking-based scoring strategy;
3) pruning data samples to increase the quality of dataset;
4) relabeling them with a ranking-based scoring strategy.

\paragraph{Collecting Synthetic Data}

Pseudo datasets for model pre-training has been proven effective for obtaining well-performed evaluation models~\cite{sellam-etal-2020-bleurt,wan-etal-2021-robleurt,wan-etal-2022-unite}.
Moreover, as in~\newcite{wan-etal-2022-unite}, training on three input formats requires massive pseudo examples.
Specifically, we first obtain parallel data from this year's WMT Translation task as the source-reference sentence pairs, and translate the source using online translation engines,~\textit{e.g.},~\texttt{Google Translate}\footnote{\href{https://translate.google.com}{\url{https://translate.google.com}}} and~\texttt{Alibaba Translate}\footnote{\href{https://translate.alibaba.com}{\url{https://translate.alibaba.com}}}, to generate the hypothesis sentence.
As discussed in~\newcite{sellam-etal-2020-bleurt}, the conventional pseudo hypotheses are usually of high translation quality.
Consequently, the dataset hardly possesses a higher level of translation quality diversity, making it difficult to train evaluation models.
We follow existing works~\cite{wan-etal-2022-unite,sellam-etal-2020-bleurt} to apply the word and span dropping strategy to attenuate hypotheses quality, increasing the ratio of training examples consisting of bad translation outputs.

\paragraph{Data Labeling and Pruning}
After downgrading the translation quality of synthetic hypothesis sentences, we then collect predicted scores for each triple as the learning supervision using checkpoint from \textsc{UniTE}~\cite{wan-etal-2022-unite}.\footnote{https://github.com/wanyu2018umac/UniTE}
As discussed in~\newcite{wan-etal-2022-unite} and~\newcite{sellam-etal-2020-bleurt}, scores labeled by low-quality metrics have poor consistency, confusing the model learning during the training period.
To increase the confidence of pseudo-labeled scores, we use multiple~\textsc{UniTE} checkpoints trained with different random seeds to label the synthetic data~\cite{wan-etal-2022-unite}.
Besides, to reduce the gap of predicted scores among different translation directions, as well as alleviate the bias among multiple evaluation approaches, we follow the scoring methods in \textsc{UniTE}~\cite{wan-etal-2022-unite}, using the idea of Borda count~\cite{ho1994decision, emerson2013original}.
After sorting the collected prediction scores, we use their ranking indexes instead, and apply the conventional Z-score strategy to normalize them.

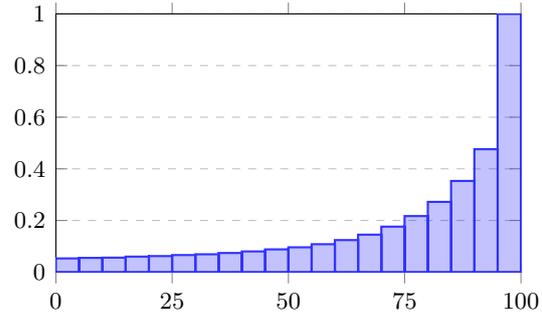
\begin{figure}[t]
    \centering
        \begin{tikzpicture}
            \pgfplotsset{set layers}
            \pgfplotsset{every x tick label/.append style={font=\small}}
            \pgfplotsset{every y tick label/.append style={font=\small}}
            
            \begin{axis}[
                ybar,
                bar width=0.04 * \columnwidth,
                height=0.65 * \columnwidth,
                width=\columnwidth,
                xmin=0.0, xmax=1.0,
                ymin=0, ymax=1.0,
                xtick={0, 0.25, 0.5, 0.75, 1.0},
                xticklabels={$0$, $25$, $50$, $75$, $100$},
                ymajorgrids=true,
                grid style=dashed,
                legend cell align=left,
                legend style={
                    at={(0.70, 1.05)},
                    anchor=south east,
                    font=\tiny,
            		legend columns=2},
            	every axis plot/.append style={thick},
                ]
            \addplot[
        	        color=blue!80!white,
        	        fill=blue!80!white,
        	        fill opacity=0.3
                ]
                plot coordinates {
                (0.025, 0.053)
                (0.075, 0.055)
                (0.125, 0.056)
                (0.175, 0.06)
                (0.225, 0.062)
                (0.275, 0.066)
                (0.325, 0.069)
                (0.375, 0.074)
                (0.425, 0.08)
                (0.475, 0.088)
                (0.525, 0.096)
                (0.575, 0.108)
                (0.625, 0.124)
                (0.675, 0.145)
                (0.725, 0.176)
                (0.775, 0.217)
                (0.825, 0.272)
                (0.875, 0.353)
                (0.925, 0.476)
                (0.975, 1.0)
            };
            \end{axis}
        \end{tikzpicture}
        \vspace{-1ex}
        \caption{The cumulative distribution of scores in WMT 2020 and 2021 MQM datasets. The x-axis represents the annotated score while the y-axis represents the ratio.}
        \label{fig.cumulative_distribution_MQM}
   
\end{figure}

During our preliminary experiments, we find that the quality of hypotheses in the MQM 2020 and 2021 dataset is generally high.
As shown in Figure~\ref{fig.cumulative_distribution_MQM}, more than 64\% of the human-annotated scores are higher than 90.
To further mitigate the disagreement of translation quality distributions between pre-training and test datasets, we arrange data pruning for synthetic data.
Specifically, for each language pair, we ascendingly sort the synthetic examples by their scores, and split the examples into 5 bins.
For the examples in each bin, we randomly drop 90\%, 80\%, 60\%, 20\%, and 0\% data examples, yielding.
We obtain 0.5M synthetic data for each language pair, and renormalize our prediction scores by the ranking-based manners as described before.
In total, we collect pseudo examples on 10 translation directions,~\textit{i.e.}, English $\leftrightarrow$ Czech/German/Japanese/Russian/Chinese, each of which contains 0.5M data tuples formatted as $\langle \mathbf{h}, \mathbf{s}, \mathbf{r}, q \rangle$.

\subsection{Training Pipeline}
\label{subsec.training_pipeline}
To train~\textsc{UniTE} models, the available datasets consist of synthetic examples (as in \S\ref{subsec.constructing_synthetic_data}), human annotations (\textit{i.e.}, DA and MQM), as well as provided development set for this year.
In practice, we arrange the training pipeline into three steps as follows.

\paragraph{Pre-train with Synthetic Data}
As illustrated in \S\ref{subsec.constructing_synthetic_data}, after collecting synthetic dataset, we use them to continuously pre-train our~\textsc{UniTE} models to enhance the evaluation ability on three input formats.

\paragraph{Fine-tune with DA Dataset}
After collecting pre-trained checkpoints, we first fine-tune them with human-annotated DA datasets.
Although the DA and MQM datasets have different scoring rules, training~\textsc{UniTE} models on DA as an additional phase can enhance both the model robustness and the support of multilinguality.
In practice, we collect all DA datasets from the year 2017 to 2020, yielding 853k training examples.
Notably, we leave the year 2021 out of training due to the reported bug from the organizational committee.

\paragraph{Fine-tune with MQM Dataset}
For the evaluation test set which is assessed with MQM scoring rules, we arrange the MQM dataset from the year 2020 and 2021 for fine-tuning models at the end of the training phase, consisting of 75k examples.
Specifically, during this step, we first use the provided development set to tune hyper-parameters for continuous pre-training and fine-tuning, and directly use all data examples to fine-tune our~\textsc{UniTE} models following the previous setting.

\subsection{Results Conduction}
To select appropriate checkpoints, we evaluate our models on this year's development set and select top-3 models for each translation direction.
Furthermore, to fully utilize the development set, we conduct a 5-fold cross-validation on the development set to select the best hyper-parameters for each top-3 model training on them.
Finally, we use the best hyper-parameters to fine-tune one single model on the entire development set.

As to the results conduction, we first applied multiple random seeds for each setting, and select the checkpoint with the best performance for model training.
Besides, to further increase the accuracy of ensembled scores, we choose two checkpoints whose backbones are XLM-R and~\textsc{InfoXLM}, respectively.

Notably, uncertainty estimation has been verified in Machine Translation and Translation Evaluation communities ~\cite{wan2020self, zhou2020uncertainty, glushkova2021uncertainty}.
However, applying this method is time consunming and we do not try it in this year's QE task. 
\begin{table*}[t]
    \centering
    \scalebox{1.0}{

        \begin{tabular}{lcccc}
            \toprule
            
        \textbf{Model} & \textbf{Multilingual} & \textbf{En-De} & \textbf{En-Ru} & \textbf{Zh-En}
            \\
             \midrule
            COMET-QE-21~\cite{zerva-etal-2021-ist} & 39.8 & 49.4 & 46.5 & 23.5 \\
            
            \cdashline{1-5}[1pt/2.5pt]\noalign{\vskip 0.5ex}
            \textsc{UniTE}-pretrain & 14.0 & 36.0 & 15.2 & 23.8 \\
            \textsc{UniTE}-pretrain-prune & 28.5 & 41.5 & 22.2 & 20.4 \\
            \textsc{UniTE}-pretrain-prune~+~DA  & \textbf{44.5} & 49.3 & 50.3 & 25.2 \\
            \textsc{UniTE}-pretrain-prune~+~MQM & 29.2 & 39.8 & 49.0 & 23.9 \\
            \textsc{UniTE}-pretrain-prune~+~DA~+~MQM & 40.2 & \textbf{52.3} & 58.5 & 25.7 \\
            \textsc{UniTE-InfoXLM}-pretrain-prune~+~DA~+~MQM & 32.2 & 47.7 & \textbf{59.0} & \textbf{27.1} \\
            
            \bottomrule
             
        \end{tabular}
    }
    \caption{Spearman's correlaion (\%) on this year's development dataset. The best result for each translation direction are bolded. Applying both DA and MQM datasets for fine-tuning can achieve better results. Taking XLM-R as backbone shows better result on En-De, and~\textsc{infoXLM} on Zh-En and En-Ru.}
    \label{table_main}
\end{table*}

\begin{table*}[t]
    \centering
    \scalebox{1.0}{

        \begin{tabular}{lcccc}
            \toprule
            
        \textbf{Model} & \textbf{Multilingual} & \textbf{En-De} & \textbf{En-Ru} &  \textbf{Zh-En}
            \\
             \midrule
            Single model & 41.1 & 46.1 & 47.4 & 31.3 \\
            \cdashline{1-5}[1pt/2.5pt]\noalign{\vskip 0.5ex}
            5-fold ensembling & 42.7 & 53.1 & 48.4 & \textbf{34.7}  \\
            XLM-R~+~\textsc{InfoXLM} ensembling & \textbf{45.6} & \textbf{55.0} & \textbf{50.5} & 33.6 \\

            \bottomrule
             
        \end{tabular}
    }
    \caption{Spearman's correlaion (\%) on this year's test set. The best results for each translation direction are viewed in bold. Using 5-fold ensembling strategy delivers better correlation on Zh-En translation direction, and ensembling models trained on different PLM backbones conducts better results on multilingual, En-De, and En-Ru setting.}
    \label{table_test}
\end{table*}

\section{Experiments}
\paragraph{Experiment Settings}
We choose the large version of XLM-R~\cite{conneau-etal-2020-unsupervised} and~\textsc{InfoXLM}~\cite{chi-etal-2021-infoxlm} as the PLM backbones of all~\textsc{UniTE} models.
The feedforward network contains three linear transition layers, whose output dimensionalities are 3,072, 1,024, and 1, respectively.
Between any two adjacent layers, a hyperbolic tangent is arranged as the activations.

During the pre-training phase, we use the WMT 2021 MQM dataset as the development set to tune the hyper-parameters for continuous pre-training and DA fine-tuning phases.
For the XLM-R setting, we apply the learning rate as $1.0 \cdot 10^{-5}$ for PLM, and $3.0 \cdot 10^{-5}$ for the feedforward network.
Especially, for~\textsc{InfoXLM} setting, we halve the corresponding learning rates to maintain the training stability.
Besides, we find that raising the batch size can make the training more stable.
In practice, we set the batch size for each input format as 1,024.
For the following fine-tuning steps, we use the batch size as 32 across all settings.

\paragraph{Evaluation Setup}
As requested by organizers, we primarily evaluate our systems in terms of Spearman's correlation metric between the predicted scores and the human annotations for each translation direction. 
Apart from that, we also take other metrics,~\textit{e.g.}, Pearson's correlation, into account.
Note that, during the evaluation of the multilingual phase, we directly calculate the correlation score for all predictions instead of conducting that for each language direction individually.

\paragraph{Baseline}
We introduce COMET-QE-21~\cite{zerva-etal-2021-ist}, one of the best-performed QE models as our strong baseline.
COMET-QE-21 have shown their strong performance in WMT 2021 QE~\cite{specia-etal-2021-findings} and Metrics Shared Task~\cite{freitag-etal-2021-results} competitions.
We directly apply the official released COMET-21-QE baseline\footnote{\href{https://github.com/Unbabel/COMET/}{\url{https://github.com/Unbabel/COMET/}}}, and use the well-trained checkpoints to infer on this year's development set for comparison.

\paragraph{Main Results}
We first testify the effectiveness of our systems on this year's development set.
As shown in Table~\ref{table_main}, our models outperform COMET-QE-21 in all translation directions.
As to the results of final submissions, we list the results in Table~\ref{table_test}.

\section{Analysis}
In this section, we discuss the effectiveness of all strategies,~\textit{i.e.}, data pruning (\S\ref{subsec.data_pruning}), training data arrangement (\S\ref{subsec.training_data}), backbone selection (\S\ref{subsec.backbone_selection}), and model ensembling methods(\S\ref{subsec.ensmeble_methods}).

\subsection{Data pruning}
\label{subsec.data_pruning}
We first investigate the impact of the data pruning strategy in Table~\ref{table_main}.
When using the pruneped data to train~\textsc{UniTE} models, the performance gains significant improvements, with 14.5, 5.5, and 7.0 Spearman's correlation on Multilingual, En-De, and En-Ru translation direction, respectively.
As discussed in \S\ref{subsec.constructing_synthetic_data}, most training examples in MQM dataset have a higher translation quality.
The data pruning method can reduce the ratio of training examples that contains poorly translated hypotheses.
In contrast to the unpruneped synthetic dataset, the ratio of those examples consisting of well-translated outputs is raised.
Consequently, we can reduce the translation quality distribution gap between synthetic and MQM datasets, and continuous pre-training and fine-tuning phases can share a great deal of learned knowledge.
The experimental results validate our thinking, that the data pruning strategy offers a higher transferability of quality evaluation from synthetic to MQM data examples, making the model learning easier on the latter.

\subsection{Training Data}
\label{subsec.training_data}
To identify which dataset among DA and MQM is more important during fine-tuning, we conduct an experiment for comparing the corresponding effectiveness.
As shown in Table~\ref{table_main}, using DA or MQM dataset can both give performance improvement compared to only using synthetic data.
Notably, the combination of DA and MQM datasets can further boost the performance in En-Ru/En-De/Zh-En directions.
However, when comparing \textsc{UniTE-DA-MQM} to \textsc{UniTE-DA}, an unexpected performance drop in the Multilingual setting is observed.

We think the reasons behind this phenomenon are two-fold.
On one hand, DA data has 34 translation directions, while MQM data only has three specific directions (\textit{i.e.}, En-De, En-Ru, and Zh-En).
The annotation rules applied for those two datasets are inconsistent with each other.
Training the model on MQM data can boost the performance in a specific direction.
While a model trained on DA data is possessed with a more general evaluation ability for more translation directions, thus delivering more stable results on multilingual evaluation scenarios.
On the other hand, for MQM data items, even though the scores may be similar across translation directions and competition years, the corresponding translation quality may vary vastly.
For example, a score of 0.3 may be relatively a high score in MQM 2021 Zh-En subset, while it is rather low in this year's En-De direction.
This phenomenon is quite critical when handling examples from multiple translation directions.
As scores from the involved two translation directions are not compatible, training on those examples concurrently may downgrade the multilingual performance of our models.

\subsection{Backbone Selection}
\label{subsec.backbone_selection}
As in Table~\ref{table_main}, UniTE-pretrain-prune~+~DA~+~MQM is trained with \textsc{XLM-R}  backbone, while \textsc{UniTE-InfoXLM}-pretrain-prune~+~DA~+~MQM is trained with \textsc{InfoXLM} using the same hyper-parameters and strategy.
As seen, after updating the backbone of~\textsc{UniTE} model from XLM-R to~\textsc{InfoXLM}, the latter model outperforms the former in En-Ru and Zh-En directions, with the improvement of Spearman's correlation at 0.5 and 1.4, respectively.
We can see that the quality estimation model can benefit from the cross-lingual alignment knowledge  during model training.
However, as to the En-De direction, the performance shows a significant drop at 4.6.
We attribute this to the reason, that English and German are from the same language family, where the two languages can obtain a great deal of cross-lingual knowledge via similar tokens with the same meaning.
For Multilingual direction, we claim that the impact of training data makes it unconfident which has been discussed in \S\ref{subsec.training_data}.

\subsection{Ensemble Methods}
\label{subsec.ensmeble_methods}
As in Table \ref{table_test}, the ensembled models show great improvement on all translation directions.
The difference between XLM-R and \textsc{InfoXLM} lies in the training objective and applied training dataset.
For the quality estimation task whose core lies in the semantic alignment across languages, the knowledge engaged inside those two PLM models can be complementary to each other.
Except for Zh-En direction, XLM~+~\textsc{InfoXLM} ensembling outperforms the 5-fold ensembling method in three tracks, with the performance increase being 2.9, 1.9, and 2.1 for Multilingual, En-De, and En-Ru settings, respectively.
This demonstrates that, ensembling models constructed with different backbones can give better results compared to the k-fold ensembling strategy.

\section{Conclusion}
In this paper, we describe our \textsc{UniTE} submission for the sentence-level MQM task at WMT 2022. 
We apply data pruning and a ranking-based scoring strategy to collect massive synthetic data.
During training, we utilize three input formats to train our models on our synthetic, DA, and MQM data sequentially.
Besides, we ensemble the two models which consist of two different backbones -- \textsc{XLM-R} and \textsc{InfoXLM}.
Experiments show that, our unified training framework can deliver reliable evaluation results on QE tasks, showing the powerful transferability of~\textsc{UniTE} model.

For future work, we believe that exploring the domain adaption problem for QE is an essential task.
The existing machine translation system has made great progress in the field of domain transferablity~\cite{lin2021towards, yao2020domain, wan-etal-2022-challenges}.
Nevertheless, the confident evaluation metrics for those translation systems are few to be explored.
Apart from that, developing a unified framework with high transferability for evaluating translation and other natural language generation tasks~\cite{yang-etal-2021-pos, yang-etal-2022-gcpg, liu2022kgr4} is quite an interesting direction.

Notably, we also participated in this year's WMT Metrics Shared Task with the same models.
We believe that, the idea of unifying three kinds of translation evaluation functionalities (\textit{i.e.}, source-only, reference-only, and source-reference-combined) into one single model can deliver dominant results on all scenarios.
Better solutions for achieving this goal are worth to be explored in the future.

\section*{Acknowledgements}

The participants would like to send great thanks to the committee and the organizers of the WMT Quality Estimation Shared Task competition.
Besides, the authors would like to thank the reviewers and meta-review for their insightful suggestions.

This work was supported in part the by the National Key Research and Development Program of China (No. 2020YFB1406703), Science and Technology Development Fund, Macau SAR (Grant No. 0101/2019/A2), the Multi-year Research Grant from the University of Macau (Grant No. MYRG2020-00054-FST), National Key Research and Development Program of China (No. 2018YFB1403202), and Alibaba Group through Alibaba Innovative Research (AIR) Program.
\bibliographystyle{acl_natbib}
\bibliography{anthology}

\end{document}